\documentclass[10pt,twocolumn,letterpaper]{article}

\usepackage[pagenumbers]{iccv} 

\usepackage{multirow}

\usepackage[table,dvipsnames]{xcolor}
\usepackage{subcaption}
\usepackage{stfloats}
\definecolor{lightgray}{gray}{0.9}
\definecolor{mediumgray}{gray}{0.8}
\definecolor{lightblue}{RGB}{200,220,255} 
\definecolor{secondblue}{RGB}{80,150,220}
\definecolor{bestred}{RGB}{220,50,50}

\definecolor{iccvblue}{rgb}{0.21,0.49,0.74}
\usepackage[pagebackref,breaklinks,colorlinks,allcolors=iccvblue]{hyperref}

\def\paperID{*****} 
\def\confName{ICCV}
\def\confYear{2025}

\title{StegGNN: Learning Graphical Representation for Image Steganography}

\author{
Abhinav Kumar \quad Shorya Singhal \quad Agam Pandey \quad Tushar Kumar \quad Sukrit Jindal\\
Indian Institute of Technology Roorkee\\
{\tt\small abhinav\_k@ma.iitr.ac.in}
}

\begin{document}
\maketitle
\begin{abstract}
Image steganography refers to embedding secret messages within cover images while maintaining imperceptibility. Recent advances in deep learning—primarily driven by Convolutional Neural Networks (CNNs) and architectures such as inverse neural networks, autoencoders, and generative adversarial networks—have led to notable progress. However, these frameworks are primarily built on CNN architectures, which treat images as regular grids and are limited by their receptive field size and a bias toward spatial locality.
In parallel, Graph Neural Networks (GNNs) have recently demonstrated strong adaptability in several computer vision tasks, achieving state-of-the-art performance with architectures such as Vision GNN (ViG). This work moves in that direction and introduces StegGNN—a novel autoencoder-based, cover-agnostic image steganography framework based on GNNs.
By modeling images as graph structures, our approach leverages the representational flexibility of GNNs over the grid-based rigidity of conventional CNNs. We conduct extensive experiments on standard benchmark datasets to evaluate visual quality and imperceptibility. Our results show that our GNN-based method performs comparably to existing CNN benchmarks. These findings suggest that GNNs provide a promising alternative representation for steganographic embedding and open the field of deep learning-based steganography to further exploration of GNN-based architectures.

\end{abstract}    
\section{Introduction}
Steganography is a widely studied technique \cite{chanu975image} that involves hiding secret data within container media such as images \citep{kadhim2019comprehensive}, videos \citep{mstafa2017compressed}, and text \citep{majeed2021review} to prevent detection by unauthorized parties. Unlike cryptography, which prioritizes robustness, steganography focuses on capacity and invisibility—ensuring that only authorized receivers can recover hidden messages while maintaining the stego medium's indistinguishability from the original. 

\begin{figure}[h]
    \hfill
    \includegraphics[height=5cm,width=8cm]{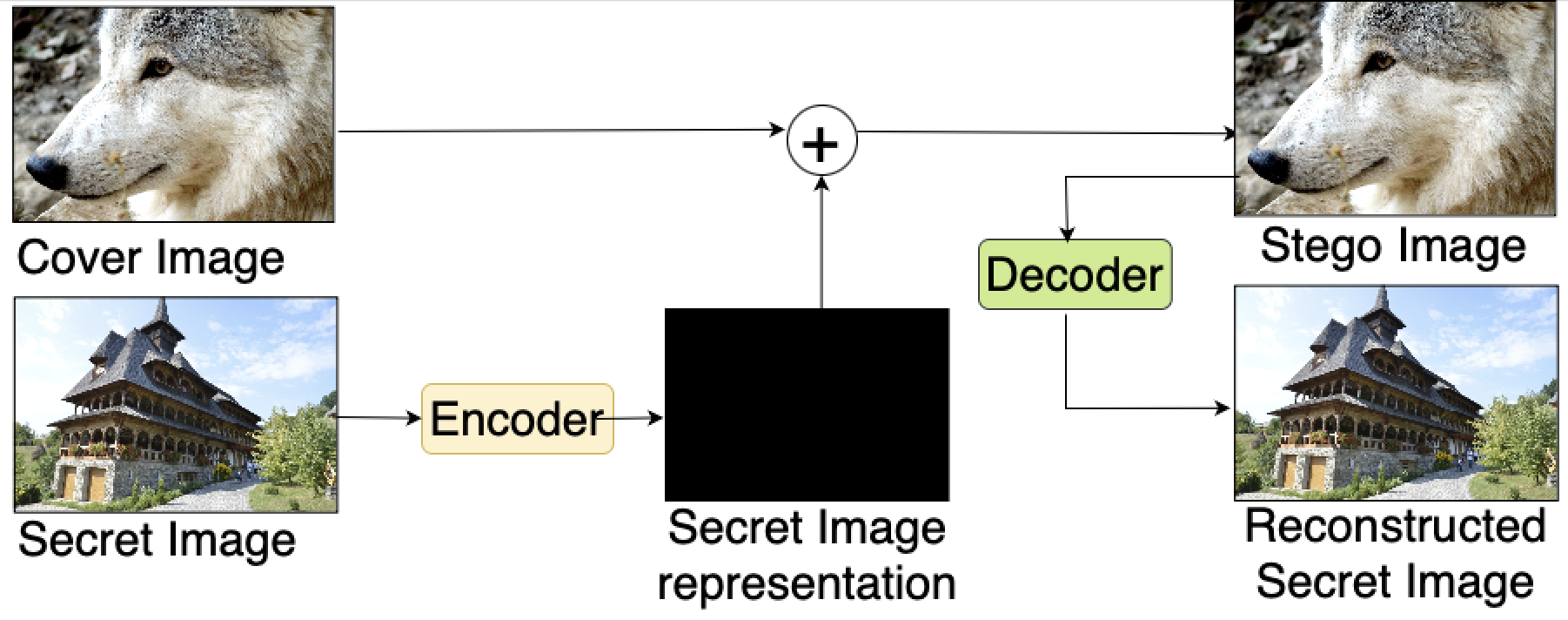}
    \caption{Overall autoencoder framework for cover agnostic steganography.}
    \label{fig:introfig}
\end{figure}

This leads to its widespread applications in copyright protection \citep{altaay2012introduction}, digital watermarking \citep{cox2007digital}, and information retrospection \citep{zhang2020viscode}. Traditional image steganography methods operate in spatial \citep{niimi2002high,kawaguchi1999principles} or frequency domains \citep{almohammad2008high}, typically embedding data in the least significant bits \citep{chan2004hiding,mielikainen2006lsb}. However, these approaches suffer from limited capacity (0.2-4 bits per pixel) and vulnerability to statistical detection \citep{fridrich2001reliable}.  With the advent of deep learning over the past decade, a surge of deep learning-based approaches \citep{song2024survey,hu2024learning} have been proposed for image steganography. 

In his seminal work, Baluja \citep{baluja2017hiding} proposed the first end-to-end CNN autoencoder framework for hiding complete images within carrier images. This was complemented by \citep{hayes2017generating}, who introduced adversarial training for steganography. Building on these foundational works, \citep{zhu2018hidden} developed the first comprehensive watermarking framework that integrated autoencoder architecture with noise layers and adversarial discriminators to enhance robustness and stego image quality. This autoencoder paradigm inspired numerous subsequent works focusing on improved imperceptibility and visual fidelity \citep{duan2019reversible,huang2022image,rahim2018end,wu2018stegnet}.  Parallel to autoencoder-based approaches, \citep{lu2021large} introduced an alternative framework based on invertible neural networks (INNs) for large-capacity steganography, spawning a series of INN-based methods \citep{xu2022robust,guan2022deepmih,jing2021hinet}. While INN-based approaches offer computational efficiency, they impose strict weight-sharing constraints between encoder-decoder architectures, limiting the architectural flexibility of autoencoder frameworks.

The above methods come under cover-dependent steganography framework \citep{zhang2020udh}. Particularly, they embed secret messages by modifying a specific cover image. The encoding is tightly coupled to the cover, thus decreasing its scalability for copyright protection and watermarking applications. In contrast, several researchers have focused on cover-less steganography to avoid altering a given cover image altogether \citep{qin2019coverless}. However, such frameworks have primarily focused on hiding bits in container images \citep{liu2020coverless}, and only a few explorations involve hiding images without resorting to cover images \citep{yu2023cross}. Furthermore, these are not generalizable across various image domains, as the generated stego image quality and diversity depend on the generative model \citep{yu2023cross}. Parallel to these approaches, \citep{zhang2020udh} proposed a cover-agnostic steganography, which decouples the secret encoding from the specific cover image, leading to universal, cover-agnostic secret embeddings. 

In this work, we adopt the cover-agnostic method, as it provides an effective and efficient way to model image data as graphs. Specifically, given a cover image $C$ and a secret image $S$, our encoder module processes $S$ to produce a universal secret embedding $S_e$. This embedding is then added to any cover image to produce the corresponding stego image $C'$, which our decoder module uses to reconstruct the original secret message $S'$ (Fig.~\ref{fig:introfig}). Compared to convolutional neural networks (CNNs)~\citep{lecun1998gradient}, which have previously been used in encoder and decoder modules~\citep{zhang2020udh}, our approach leverages recent advances in graph neural networks (GNNs) for vision domains~\citep{han2022vision, munir2023mobilevig, munir2024greedyvig, spadaro2024wignet} to capture long-range dependencies.

\section{Related Work}
\label{sec:formatting}

In the following subsection, we briefly discuss the progress made by cover-dependent and cover-independent steganography methods, as well as Graph Neural Networks.

\subsection{Cover Dependent} 

Cover-dependent steganography modifies an existing cover image to embed secrets. Early spatial-domain techniques—such as least significant bit (LSB) replacement \citep{chan2004hiding,mielikainen2006lsb} offered simple embedding but were vulnerable to statistical steganalysis \citep{imaizumi2014multibit,provos2003hide}. Frequency-domain approaches (e.g. DFT \citep{ruanaidh1996phase}, DCT \citep{hsu1999hidden} and DWT \citep{barni2001improved}) improved robustness and undetectability, but were limited to bit-level payloads and were unsuitable for hiding full images. To overcome these constraints, deep neural network (DNN)–based steganography was proposed in \citep{baluja2017hiding} which spurred a series of advancements, support for embedding multiple hidden images \citep{baluja2019hiding}, techniques for concealing binary data, text and watermarks \citep{tancik2020stegastamp, zhu2018hidden, wengrowski2019light}. Concurrent research leveraging invertible neural networks (INNs) \citep{lu2021large,dinh2014nice,xiao2020invertible} framed the tasks of embedding and recovery as inverse problems, enabling the encoder and decoder to share weights within a unified architecture and inspired subsequent efforts \citep{xu2022robust,ye2024pprsteg,jing2021hinet}. Recent improvements leverage generative adversarial networks (GANs) to further improve the perceptual quality of stego-images \citep{zhang2019steganogan,hayes2017generating}.

\subsection{Cover Independent}  

Cover-independent (or agnostic) steganography decouples the secret encoding process entirely from the cover image \citep{zhang2020udh}. Unlike cover-dependent methods that bind the secret with unique cover features, cover-independent techniques generate a universal secret representation. By leaving the cover image unmodified, this approach preserves its inherent statistical properties and visual quality, avoiding artifacts introduced by synthetic stego generation as seen in some coverless methods. Moreover, the same secret can be embedded across diverse covers, enabling universal watermarking and light field messaging applications \citep{zhang2020udh}.

\subsection{Graph Neural Networks}

GNNs emerged as an extension of the convolution operation of CNNs for regular-structured data such as images to the graph domain. \citep{bruna2014spectral} proposed the first modern GNN by extending the convolutional operator of CNNs \citep{lecun1998gradient} to graphs. Incorporating concepts of signal processing on graphs \citep{defferrard2016convolutional}, introduced localized spectral filtering for graphs. Later \citep{kipf2017semi}, approximated the spectral filtering
operation to obtain efficient Graph Convolutional Networks (GCNs). GNNs are typically used for learning graph-structured data representations such as social networks \citep{hamilton2017inductive}, citation networks \citep{sen2008collective}, and biochemical graphs \citep{wale2008comparison}. Recently, Graph Convolution Networks (GCNs)/GNNs have been introduced in computer vision domains where they have been used for scene graph generation \citep{xu2017scene}, image segmentation \citep{giraldo2022hypergraph}, and classification \citep{spadaro2024wignet,han2022vision}. Moving in this direction, this paper is the first to extend GNNs/GCNs in image steganography.

\begin{figure}[h]
    \hfill
    \includegraphics[height=5cm,width=8cm]{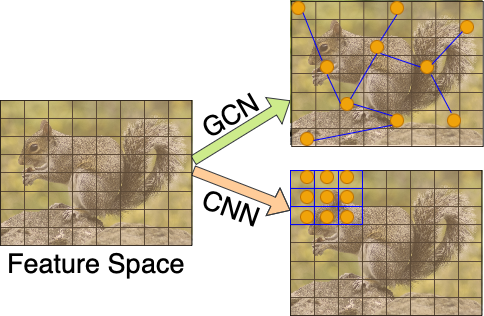}
    \caption{Difference between Graph Convolution Network(GCN) and Convolution Neural Network(CNN) operation.}
    \label{fig:right_figure_single}
\end{figure}

\begin{figure*}[t]
    \centering
    \includegraphics[height=7cm,width=18.5cm]{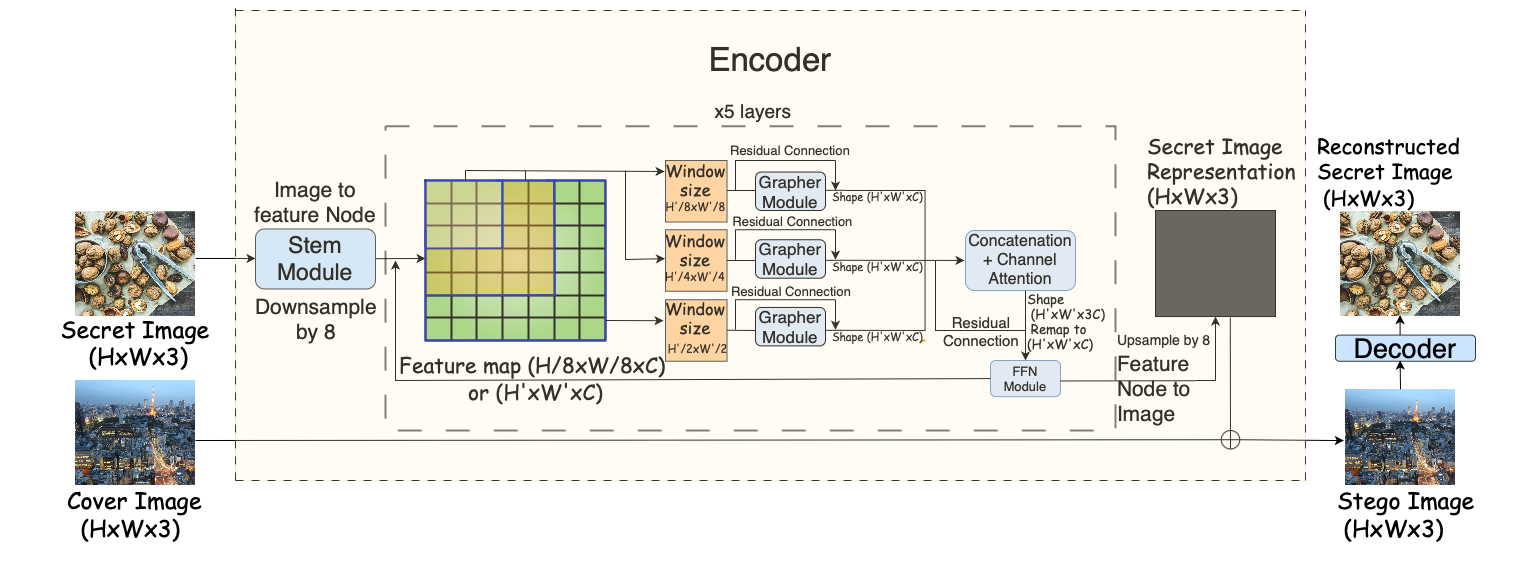}
    \caption{Overall framework of the proposed StegGNN model, illustrating the different components of our pipeline.}
    \label{fig:method}
\end{figure*}

\section{Methodology}

\subsection{Overview}

We adopt a cover-independent autoencoder framework, as proposed by \citep{zhang2020udh}. Specifically, given a secret image $\mathbf{S}$ and a cover image $\mathbf{C}$, the encoder module processes the secret image and generates a universal secret representation $\mathbf{S_e}$, which can be added to any cover image to produce the stego image $\mathbf{C'}$. The decoder module then takes the stego image as input and reconstructs the original secret image, yielding $\mathbf{S'}$ (Fig.~\ref{fig:method}). For simplicity, we use similar architectures for both the encoder and decoder modules. Following previous work \citep{baluja2017hiding,zhang2020udh}, we train the entire network using the following loss, where $||\cdot||$ denotes the $\ell_2$ norm and $\beta$ is set to $0.75$.

\[
\mathcal{L}(\mathbf{C}, \mathbf{C'}, \mathbf{S}, \mathbf{S'}) = \|\mathbf{C} - \mathbf{C'}\| + \beta \|\mathbf{S} - \mathbf{S'}\|
\]

\subsection{Encoder Module Architecture}

\begin{figure*}[h]
    \centering
    \includegraphics[height=4.5cm,width=16cm]{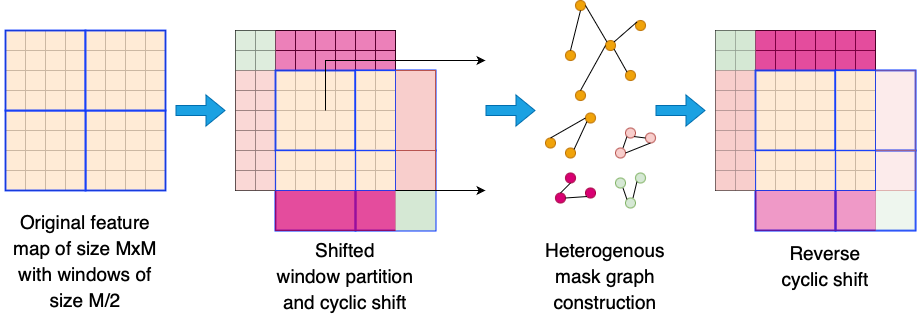}
    \caption{Illustration of the cyclic shifting process used to obtain shifted windows for cross window interaction}
    \label{fig:shifted}
\end{figure*}

\subsubsection{Image to Feature Node}
\paragraph{Stem Module.} Given a secret image of size $H \times W \times 3$, the encoder module first divides the image into overlapping patches \citep{wang2022pvt}. Specifically, we apply a convolutional operation with stride $S$, kernel size $2S - 1$, and padding size $S - 1$, producing an output tensor of dimensions $\frac{H}{S} \times \frac{W}{S} \times C_h$, where $C_h$ is the number of output channels. Motivated by recent works \citep{islammuch, chuconditional}, we discard the fixed-size positional encoding \citep{dosovitskiy2020image} in favor of zero-padding-based positional encoding during patchification. Consequently, we obtain $\mathit{N} = \frac{H}{8} \times \frac{W}{8}$ nodes, corresponding to three consecutive downsampling operations with stride 2 each.

Each overlapping patch is transformed into a feature vector $\mathbf{x}_i \in \mathbb{R}^D$, forming the matrix $\mathbf{X} = [\mathbf{x}_1, \mathbf{x}_2, \dots, \mathbf{x}_N]$, where $D$ denotes the feature dimension and $i = 1, 2, \dots, N$. These features are treated as an unordered set of nodes $\mathcal{V} = {v_1, v_2, \dots, v_N}$ in a graph. For each node $v_i$, we identify its $K$ nearest neighbors $\mathcal{N}(v_i)$ in the feature space and establish directed edges $e{_{ji}}$ from each neighbor $v_j \in \mathcal{N}(v_i)$ to $v_i$, as described in detailed in the next section.

\subsubsection{Shifted Window-based Grapher Module}

Given the feature matrix $\mathbf{X} \in \mathbb{R}^{N \times D}$ produced by the Stem module (or previous Grapher layers), we apply a \textit{window partitioning} operation \citep{liu2021swin} that divides the spatial representation into non-overlapping windows of fixed spatial size $M \times M$. To capture both short- and long-range dependencies effectively, we perform multiscale processing by partitioning and processing the input tensor \(\mathbf{X}\) in parallel using windows of three different sizes: \( M \in \{\left\lfloor\frac{H^{'}}{2}\right\rfloor, \left\lfloor\frac{H^{'}}{4}\right\rfloor, \left\lfloor\frac{H^{'}}{8}\right\rfloor\} \). This way, larger windows effectively model the global context, while smaller ones focus on local patterns.

Further, we incorporate a shifted window mechanism as proposed by  \citep{spadaro2024wignet} to enable cross-window interaction between adjacent windows and avoid isolated processing. Specifically, for each window size $M$, we apply a cyclic shift of size $\left\lfloor\frac{M}{2}\right\rfloor$ along both height and width dimensions. This shift operation allows graph construction and message passing across window boundaries, promoting cross-window information flow.

To allow connections between adjacent nodes in the feature map and prevent invalid edges, we follow \citep{spadaro2024wignet} and apply a masking strategy within each shifted window that restricts connections to nodes within the same visible sub-region (Fig.~\ref{fig:shifted}). This may result in multiple disconnected subgraphs per window, with the number of valid neighbors varying across positions due to the shift. To compensate for this uneven connectivity, we scale the number of neighbors per node based on the ratio of available connections, using \( k_i = k \times \frac{P_i}{M^2} \), where \( P_i \) is the number of permissible neighbors for node \( i \). This adjustment ensures balanced and meaningful message passing across all windows.

The graph construction and convolution operations are applied independently within each shifted window, enabling localized modeling and cross-region communication. Finally, the processed features by all graph convolution layers are concatenated and passed through a channel attention module to explicitly reweight the importance of different processing and ensure the secret image is hidden across each channel effectively.

\subsubsection{Graph Neural Network and Feed Forward Network Module}

Consider a particular window $w$ consisting of a set of nodes $\mathcal{V}^{(w)}$. Within this window, we dynamically construct a directed graph $\mathcal{G}^{(w)} = (\mathcal{V}^{(w)}, \mathcal{E}^{(w)})$ by connecting each node to its $k$ nearest neighbors in the feature space using the $k$-nearest neighbor (k-NN) algorithm. This graph is reconstructed after every layer processing, allowing the edge connections to adapt based on updated node features. Importantly, the similarity between nodes is measured via dot product in the feature space rather than spatial proximity.

To perform graph-based message passing, we employ the EdgeConv operation \citep{wang2019dynamic} for feature aggregation and update. For a node $i$ in window $w$, the updated feature $\mathbf{x}_i^{\prime (w)}$ is computed as:

\begin{equation}
\mathbf{x_i}^{\prime (w)} = \max_{j \in \mathcal{N}_i^{w}} \mathbf{W}{\text{update}} \left( \left[ \mathbf{x}_i^{(w)} \mathbin{|} (\mathbf{x}_j^{(w)} - \mathbf{x}_i^{(w)}) \right] \right),
\end{equation}

where:
\begin{itemize}
\item $\mathcal{N}i^{(w)}$ denotes the set of $k$ nearest neighbors of node $i$ in window $w$,
\item $\mathbin{|}$ denotes vector concatenation,
\item $\max$ denotes channel-wise max across feature channels,
\item $\mathbf{W}{\text{update}}$ is a learnable weight matrix also known as Multi-Layer Perceptron (MLP).
\end{itemize}

\begin{figure}[h]
    \hspace*{-0.25cm}
    \includegraphics[width=0.48\textwidth]{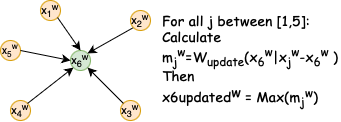}
    \caption{Illustrative example of the dynamic graph convolution
of StegGNN.}
    \label{fig:left_only}
\end{figure}

\begin{figure}[!h]
    \hspace*{-1.0cm}
    \includegraphics[height=5cm,width=9cm]{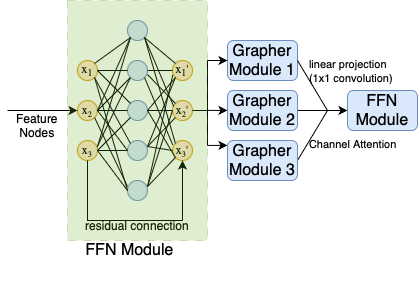}
    \caption{Illustrative example of Feed Forward Network(FFN) Module}
    \label{fig:FFN}
\end{figure}

GCNs typically suffer from over-smoothing \citep{giraldo2023trade}, where increasing depth leads to indistinguishable node features and a degradation in representational power. To address this, we introduce a Feed Forward Network (FFN) module \citep{han2022vision} following each Grapher module to increase non-linearity and enhance feature expressiveness. Additionally, to promote feature diversity and ensure consistent dimensionality before and after the graph convolution, we apply ${1\times 1}$ after channel attention reweighting (Fig.~\ref{fig:FFN}). This combination of dynamic graph construction, EdgeConv-based message passing, and FFN-based transformation allows the network to effectively capture local and global feature dependencies while maintaining high discriminative power.

\subsubsection{Inter Block Connection}

Following the initial patch embedding, the feature map is processed by a cascade of five identical \textit{Block modules}, each containing a Grapher, a FFN, and a channel-attention submodule. Let \(\mathbf{x}_0\) denote the stem output. The network maintains a running sum initialized as:
\[
  \mathrm{acc}_0 = \mathbf{x}_0,
\]
and iteratively computes for each block \(i = 1,\dots,5\):
\[
  \mathbf{y}_i = \mathrm{Block}_i(\mathrm{acc}_{i-1}),
  \quad
  \mathrm{acc}_i = \mathrm{acc}_{i-1} + \mathbf{y}_i.
\]

This additive “dense + residual” connectivity ensures that each block has access to all preceding features while preserving a local residual path \citep{zhang2018residual}. The final fused feature, \(\mathrm{acc}_5\), is then forwarded to the upsampling head

\subsubsection{Feature Node to Image}
The updated node features are reshaped and successively upsampled to convert the nodes into a secret image representation. Specifically, this involves a series of ConvTranspose2D layers \citep{zeiler2010deconvolutional} and PReLU activation \citep{he2015delving}, followed by a Tanh activation function in last. The resulting secret image representation is then added to the cover image to produce the stego image. 

The decoder follows a similar architecture, with the key difference being that it takes the stego image as input and reconstructs the secret image as output.

\begin{table*}[b]
\centering
\caption{Performance comparisons on different datasets. ``$\uparrow$'': the larger the better, ``$\downarrow$'': the smaller the better.}
\label{tab:performance}
\small
\setlength{\tabcolsep}{3pt}

\begin{tabular}{|l|ccc|ccc|ccc|}
\hline
\multicolumn{1}{|c|}{\multirow{2}{*}{Methods}} 
& \multicolumn{9}{c|}{\cellcolor{mediumgray}\textbf{Cover/Stego-image pair}} \\ \cline{2-10}
\multicolumn{1}{|c|}{} 
& \multicolumn{3}{c|}{\cellcolor{lightgray}DIV2K} 
& \multicolumn{3}{c|}{\cellcolor{lightgray}COCO} 
& \multicolumn{3}{c|}{\cellcolor{lightgray}ImageNet} \\ 
\cline{2-10}
& \cellcolor{lightgray}PSNR(dB)$\uparrow$ 
& \cellcolor{lightgray}SSIM$\uparrow$ 
& \cellcolor{lightgray}LPIPS$\downarrow$ 
& \cellcolor{lightgray}PSNR(dB)$\uparrow$ 
& \cellcolor{lightgray}SSIM$\uparrow$ 
& \cellcolor{lightgray}LPIPS$\downarrow$ 
& \cellcolor{lightgray}PSNR(dB)$\uparrow$ 
& \cellcolor{lightgray}SSIM$\uparrow$ 
& \cellcolor{lightgray}LPIPS$\downarrow$ \\ 
\hline

HIDDeN \citep{zhu2018hidden} 
& 28.49 & 0.9300 & 0.1148 
& 29.02 & 0.9283 & 0.1223 
& 28.35 & 0.9217 & 0.1455 \\

Baluja \citep{baluja2017hiding}
& 28.33 & 0.9363 & 0.1232 
& 28.95 & 0.9328 & 0.1266 
& 28.23 & 0.9242 & 0.1516 \\

UDH \citep{zhang2020udh} 
& 34.21 & 0.9419 & 0.0147 
& 35.00 & 0.9345 & 0.0164 
& 34.95 & 0.9311 & 0.0149 \\

HiNet \citep{jing2021hinet} 
& \textcolor{bestred}{\textbf{43.00}} & \textcolor{bestred}{\textbf{0.9913}} & \textcolor{bestred}{\textbf{0.0002}} 
& \textcolor{bestred}{\textbf{44.05}} & \textcolor{bestred}{\textbf{0.9906}} & \textcolor{bestred}{\textbf{0.0002}} 
& \textcolor{bestred}{\textbf{43.77}} & \textcolor{bestred}{\textbf{0.9898}} & \textcolor{bestred}{\textbf{0.0003}} \\

PUSNet \citep{li2024purified}
& 38.15 & 0.9581 & 0.0004 
& 39.09 & 0.9572 & 0.0003 
& 38.94 & 0.9456 & 0.0004 \\

\rowcolor{lightblue}
\textbf{Ours} 
& \textcolor{secondblue}{\textbf{39.55}} & \textcolor{secondblue}{\textbf{0.9682}} & \textcolor{secondblue}{\textbf{0.0003}} 
& \textcolor{secondblue}{\textbf{40.08}} & \textcolor{secondblue}{\textbf{0.9642}} & \textcolor{secondblue}{\textbf{0.0001}} 
& \textcolor{secondblue}{\textbf{40.46}} & \textcolor{secondblue}{\textbf{0.9671}} & \textcolor{secondblue}{\textbf{0.0002}} \\

\hline
\end{tabular}

\vspace{0.5cm}

\begin{tabular}{|l|ccc|ccc|ccc|}
\hline
\multicolumn{1}{|c|}{\multirow{2}{*}{Methods}} 
& \multicolumn{9}{c|}{\cellcolor{mediumgray}\textbf{Secret/Recovered image pair}} \\ \cline{2-10}
\multicolumn{1}{|c|}{} 
& \multicolumn{3}{c|}{\cellcolor{lightgray}DIV2K} 
& \multicolumn{3}{c|}{\cellcolor{lightgray}COCO} 
& \multicolumn{3}{c|}{\cellcolor{lightgray}ImageNet} \\ 
\cline{2-10}
& \cellcolor{lightgray}PSNR(dB)$\uparrow$ 
& \cellcolor{lightgray}SSIM$\uparrow$ 
& \cellcolor{lightgray}LPIPS$\downarrow$ 
& \cellcolor{lightgray}PSNR(dB)$\uparrow$ 
& \cellcolor{lightgray}SSIM$\uparrow$ 
& \cellcolor{lightgray}LPIPS$\downarrow$ 
& \cellcolor{lightgray}PSNR(dB)$\uparrow$ 
& \cellcolor{lightgray}SSIM$\uparrow$ 
& \cellcolor{lightgray}LPIPS$\downarrow$ \\ 
\hline

HIDDeN \citep{zhu2018hidden} 
& 27.66 & 0.8657 & 0.1260 
& 28.38 & 0.8636 & 0.1281 
& 28.15 & 0.8580 & 0.1413 \\

Baluja \citep{baluja2017hiding}
& 28.23 & 0.9055 & 0.1505 
& 27.43 & 0.9064 & 0.1461 
& 27.16 & 0.8957 & 0.1719 \\

UDH \citep{zhang2020udh} 
& \textcolor{secondblue}{\textbf{33.02}} & \textcolor{secondblue}{\textbf{0.9365}} & \textcolor{secondblue}{\textbf{0.0420}}
& \textcolor{secondblue}{\textbf{33.82}} & \textcolor{secondblue}{\textbf{0.9302}} & \textcolor{secondblue}{\textbf{0.0420}}
& \textcolor{secondblue}{\textbf{33.99}} & \textcolor{secondblue}{\textbf{0.9279}} & \textcolor{secondblue}{\textbf{0.0518 }} \\

HiNet \citep{jing2021hinet} 
& \textcolor{bestred}{\textbf{34.67}} & \textcolor{bestred}{\textbf{0.9655}} & \textcolor{bestred}{\textbf{0.0010}} 
& \textcolor{bestred}{\textbf{34.44}} & \textcolor{bestred}{\textbf{0.9722}} & \textcolor{bestred}{\textbf{0.0008}} 
& \textcolor{bestred}{\textbf{34.95}} & \textcolor{bestred}{\textbf{0.9647}} & \textcolor{bestred}{\textbf{0.0011}} \\

PUSNet \citep{li2024purified}
& 26.22 & 0.8363 & 0.1806 
& 26.16 & 0.8211 & 0.1710 
& 26.28 & 0.8028 & 0.1808 \\

\rowcolor{lightblue}
\textbf{Ours} 
& 28.49 & 0.8924 & 0.0651 
& 29.43 & 0.8964 & 0.0601 
& 29.04 & 0.8769 & 0.0777 \\

\hline
\end{tabular}

\end{table*}

\section{Experiments and Results}

\subsection{Training}

Our \textit{StegGNN} model is trained on the DIV2K training dataset \citep{agustsson2017ntire}, which comprises 800 high-resolution images. During training, we randomly crop $256 \times 256$ patches from the images and apply horizontal and vertical flipping as data augmentation techniques. A batch size of 8 is used, with half of the patches randomly designated as cover images and the other half as secret images. The model is trained for 3000 epochs using the Adam optimizer \citep{kingma2014adam}. The initial learning rate is set to $1 \times 10^{-4}$ and is reduced by a factor of 2 every 500 epochs.

\subsection{Evaluation and Benchmarks}

To verify the effectiveness of our approach, we evaluate our method on three datasets: the DIV2K validation dataset \citep{agustsson2017ntire}, 1000 cover-secret image pairs randomly sampled from both ImageNet \citep{russakovsky2015imagenet} and COCO \citep{lin2014microsoft}. All testing images are resized using bilinear interpolation to ensure that the cover and secret images have the same resolution of $256 \times 256$.

We compare our method against several state-of-the-art (SOTA) image-in-image hiding methods, including cover-dependent methods like Baluja \citep{baluja2017hiding}, HiDDeN \citep{zhu2018hidden}, HiNet \citep{jing2021hinet}, and PuSNet \citep{li2024purified}, and  cover-independent method like UDH \citep{zhang2020udh}. For fair comparison, we retrain all aforementioned models on the DIV2K training dataset and evaluate their performance under the same conditions as our method.

\subsection{Visual Quality}

To evaluate the visual quality of cover/stego and secret/reconstructed-secret image pairs, we adopt the following metrics: Peak Signal-to-Noise Ratio (PSNR), Structural Similarity Index Measure (SSIM~\citep{wang2004image}), and Learned Perceptual Image Patch Similarity (LPIPS~\citep{zhang2018unreasonable}).

As shown in Table~\ref{tab:performance}, our \textit{StegGNN} model outperforms HiDDeN, Baluja, UDH, and PUSNet across all three metrics and datasets in terms of stego‐image visual quality. Specifically, StegGNN achieves a performance gain of 10 dB on every dataset compared with HiDDeN and Baluja, and surpasses UDH and PUSNet by over 5 dB and 1 dB, respectively—ranking second only to HiNet. However, our decoder module does not perform as strongly as the encoder, attaining a PSNR of approximately 29 dB. Nevertheless, this is sufficient to reveal meaningful content in the reconstructed images and remains comparable with most state‑of‑the‑art baselines (HiDDeN: 28 dB; PUSNet: 26 dB; Baluja: 27 dB).

\begin{figure*}[t]
    \centering
    \includegraphics[width=0.9\linewidth]{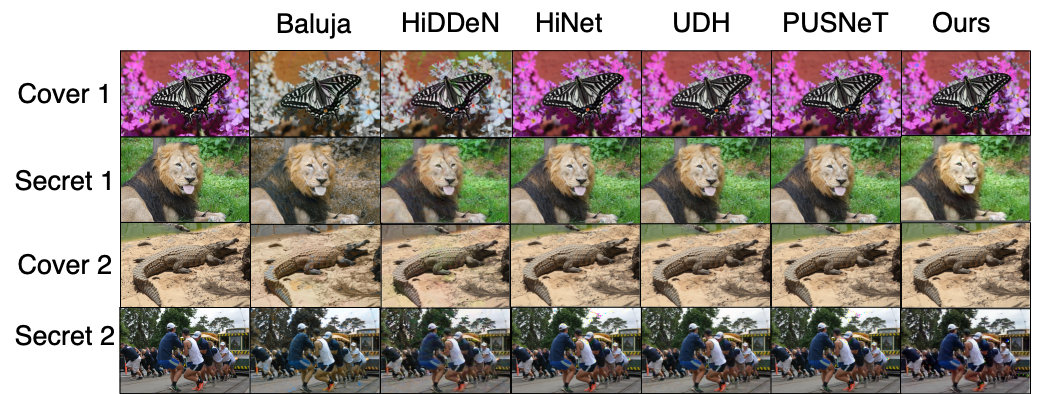}
    \caption{Examples of stego and reconstructed secret image generated by different architectures.}
    \label{fig:mydiagram}
\end{figure*}

\begin{figure*}[b]
  \centering
  \includegraphics[height=7.5cm,width=14cm]{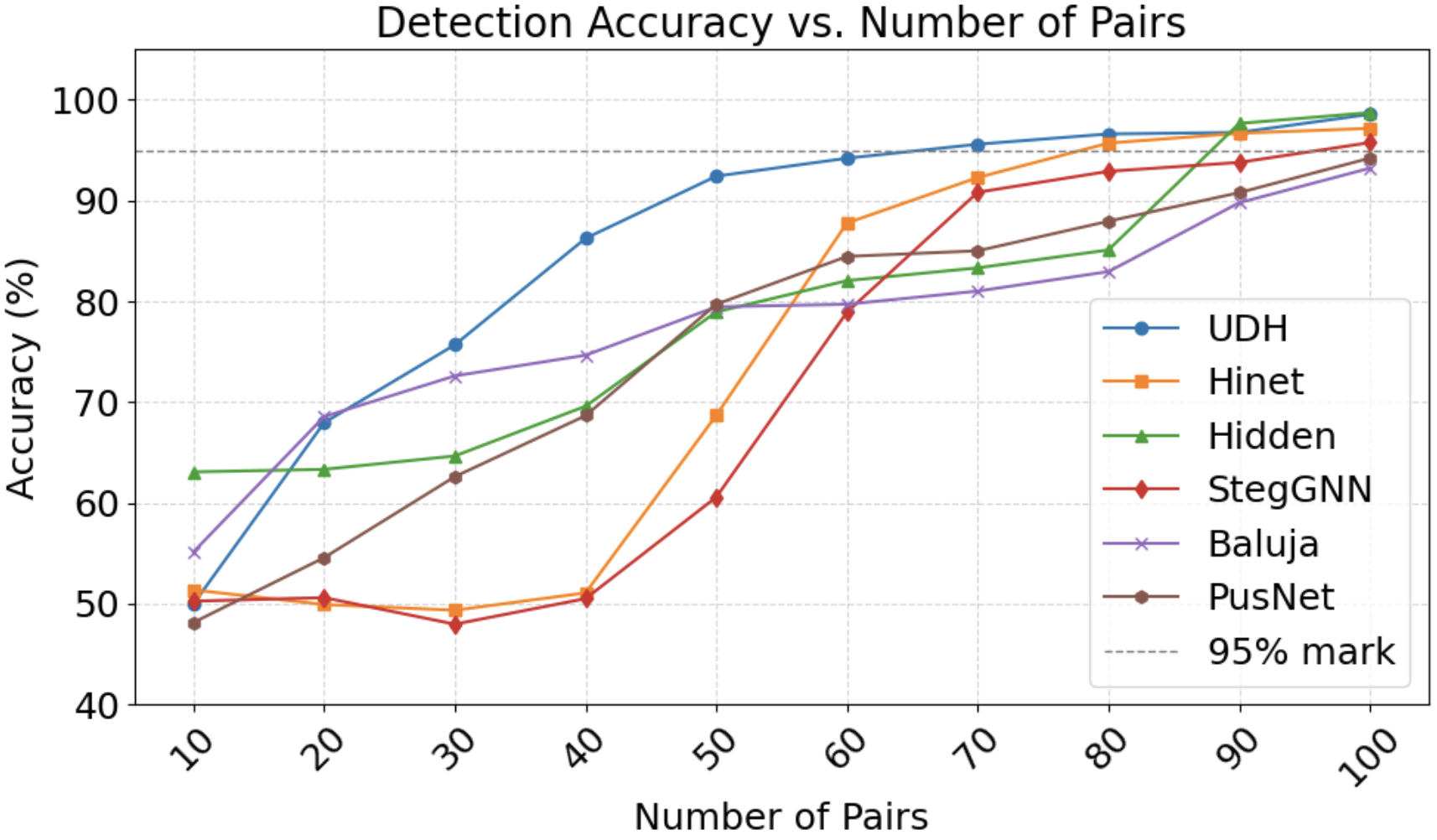}
  \caption{Detection accuracy vs.\ number of training pairs for SRNet.}
  \label{fig:srnet}
\end{figure*}

Fig.~\ref{fig:mydiagram} illustrates the stego and reconstructed secret images generated by different models. It can be observed that our StegGNN model produces high visual quality stego and reconstructed secret images that closely resemble the original cover and secret images, whereas those generated by HiDDeN and Baluja exhibit subtle visual differences.

\subsection{Steganographic analysis}

\subsubsection{Statistical steganalysis}

To evaluate the undetectability of the stego-images generated by our model against statistical methods, we follow the same protocol as proposed in~\citep{jing2021hinet,baluja2017hiding,li2024purified}. Specifically, we use an open-source steganalysis tool called \textit{StegExpose}~\citep{boehm2014stegexpose}.
We ran our StegGNN model on all cover-secret image pairs present in the three test datasets. We then fed these into StegExpose for evaluation. By varying the detection thresholds in StegExpose, we obtain a receiver operating characteristic (ROC) curve. The area under the curve (AUC) of the ROC is 0.56, which is close to random guessing (AUC = 0.5) and comparable to the results reported in PUSNet (AUC = 0.58) and HiNet (AUC = 0.51). This demonstrates the high undetectability of our stego-images against StegExpose and confirms with previous work~\citep{baluja2017hiding} that our deep learning model does not simply hide the secret image in the least significant bits of the cover image.

\subsubsection{Deep learning-based steganalysis}
To evaluate the undetectability of our stego-images in more realistic scenarios, we assessed them against popular state-of-the-art open-source deep learning-based steganalysis tools, specifically \textit{SRNet}~\citep{boroumand2018deep, jing2021hinet}. To measure the resistance of our model to detection by SRNet, we followed the protocol used in~\citep{jing2021hinet}. Specifically, we retrained the steganalysis model using varying numbers of cover/stego-image pairs to investigate how many samples are required for accurate detection. Fig.~\ref{fig:srnet} shows how the detection accuracy varies with the number of training images, where we gradually increased the amount of training data used to retrain SRNet. As the figure shows, our method outperforms cover-independent frameworks by a large margin and performs comparably with cover-dependent frameworks. Fig.~\ref{fig:srnet} indicates that an adversary requires at least 100 cover-stego image pairs to reliably detect the presence of secret data ($>$95\% accuracy). This presents a challenge in real-world applications. Since there is always a trade-off between payload capacity and undetectability~\citep{zhu2021destroying,zhu2022image}, the sender can reduce the amount of secret information embedded in the stego-image to improve its undetectability.

\section{Limitations and Future Work}

While StegGNN demonstrates the viability of graph-based cover-agnostic steganography, its decoder module achieves moderate reconstruction quality (around 29 dB PSNR), limiting fine-detail recovery.

Here we list some possible solutions which might be worth exploring in future:
\begin{enumerate}
  \item Unlike UDH, which uses a UNet-style encoder-decoder with skip connections, or HiNet, which adopts an invertible neural network (INN) for tight encoder–decoder coupling, our simple autoencoder design constrains the decoder’s expressive power. Future work could explore replacing or augmenting our decoder with a UNet-like architecture for richer multi-scale feature fusion, or adopting an INN-based framework to enforce invertibility and improve secret extraction fidelity.
  
  \item Similarly, although StegGNN resists statistical and deep-learning detectors, further hardening could be achieved by incorporating adversarial training and domain-specific losses. For example, HiDDeN employs GAN-based objectives to strengthen anti-steganalysis robustness, while HiNet uses wavelet-domain loss to penalize detectable artifacts. Integrating adversarial and wavelet losses into our training may boost undetectability without sacrificing payload capacity.
  
  \item Finally, advanced graph architectures, such as transformer-augmented GNN layers \citep{yang2021graphformers}, attention-based message passing \citep{velivckovic2018graph}, or hierarchical graph pooling \citep{bianchi2023expressive}, could be integrated to capture complex dependencies and improve overall performance. These enhancements may enable StegGNN to model higher-order relationships, close the gap with state-of-the-art CNN and INN-based steganography, and provide more flexible embedding methods.
\end{enumerate}

\section{Conclusion}

We introduce StegGNN, the first cover‑agnostic steganography framework underpinned by Graph Neural Networks. By representing image patches as graph nodes and utilizing shifted‑window $k$‑nearest‑neighbor graphs, StegGNN achieves high stego‑image visual quality and strong imperceptibility while reconstructing secret images with satisfactory details. These results are attained with a concise autoencoder architecture in a challenging cover‑independent setting. Our findings demonstrate the viability of GNNs as an effective alternative to traditional grid‑based methods and provide a foundation for future developments in steganographic model design. Our implementation is publicly available at \url{https://github.com/1609abhi/StegGNN}.

\clearpage
{
    \small
    \bibliographystyle{ieeenat_fullname}
    \bibliography{main}

@article{kadhim2019comprehensive,
  title={Comprehensive survey of image steganography: Techniques, Evaluations, and trends in future research},
  author={Kadhim, Inas Jawad and Premaratne, Prashan and Vial, Peter James and Halloran, Brendan},
  journal={Neurocomputing},
  volume={335},
  pages={299--326},
  year={2019},
  publisher={Elsevier}
}

@article{mstafa2017compressed,
  title={Compressed and raw video steganography techniques: a comprehensive survey and analysis},
  author={Mstafa, Ramadhan J and Elleithy, Khaled M},
  journal={Multimedia Tools and Applications},
  volume={76},
  pages={21749--21786},
  year={2017},
  publisher={Springer}
}

@article{majeed2021review,
  title={A review on text steganography techniques},
  author={Majeed, Mohammed Abdul and Sulaiman, Rossilawati and Shukur, Zarina and Hasan, Mohammad Kamrul},
  journal={Mathematics},
  volume={9},
  number={21},
  pages={2829},
  year={2021},
  publisher={MDPI}
}

@book{cox2007digital,
  title={Digital watermarking and steganography},
  author={Cox, Ingemar and Miller, Matthew and Bloom, Jeffrey and Fridrich, Jessica and Kalker, Ton},
  year={2007},
  publisher={Morgan kaufmann}
}

@inproceedings{altaay2012introduction,
  title={An introduction to image steganography techniques},
  author={Altaay, Alaa A Jabbar and Sahib, Shahrin Bin and Zamani, Mazdak},
  booktitle={2012 international conference on advanced computer science applications and technologies (ACSAT)},
  pages={122--126},
  year={2012},
  organization={IEEE}
}

@article{zhang2020viscode,
  title={Viscode: Embedding information in visualization images using encoder-decoder network},
  author={Zhang, Peiying and Li, Chenhui and Wang, Changbo},
  journal={IEEE Transactions on Visualization and Computer Graphics},
  volume={27},
  number={2},
  pages={326--336},
  year={2020},
  publisher={IEEE}
}

@article{chanu975image,
  title={Image Steganography and Steganalysis: A Survey},
  author={Chanu, Yambem Jina and Singh, Kh Manglem and Tuithung, Themrichon},
  journal={International Journal of Computer Applications},
  volume={975},
  pages={8887}
}

@inproceedings{almohammad2008high,
  title={High capacity steganographic method based upon JPEG},
  author={Almohammad, Adel and Hierons, Robert M and Ghinea, Gheorghita},
  booktitle={2008 Third International Conference on Availability, Reliability and Security},
  pages={544--549},
  year={2008},
  organization={IEEE}
}

@inproceedings{niimi2002high,
  title={High capacity and secure digital steganography to palette-based images},
  author={Niimi, Michiharu and Noda, Hideki and Kawaguchi, Eiji and Eason, Richard O},
  booktitle={Proceedings. International conference on image processing},
  volume={2},
  pages={II--II},
  year={2002},
  organization={IEEE}
}

@inproceedings{kawaguchi1999principles,
  title={Principles and applications of BPCS steganography},
  author={Kawaguchi, Eiji and Eason, Richard O},
  booktitle={Multimedia systems and applications},
  volume={3528},
  pages={464--473},
  year={1999},
  organization={SPIE}
}

@article{chan2004hiding,
  title={Hiding data in images by simple LSB substitution},
  author={Chan, Chi-Kwong and Cheng, Lee-Ming},
  journal={Pattern recognition},
  volume={37},
  number={3},
  pages={469--474},
  year={2004},
  publisher={Elsevier}
}

@article{mielikainen2006lsb,
  title={LSB matching revisited},
  author={Mielikainen, Jarno},
  journal={IEEE signal processing letters},
  volume={13},
  number={5},
  pages={285--287},
  year={2006},
  publisher={IEEE}
}

@inproceedings{fridrich2001reliable,
  title={Reliable detection of LSB steganography in color and grayscale images},
  author={Fridrich, Jessica and Goljan, Miroslav and Du, Rui},
  booktitle={Proceedings of the 2001 workshop on Multimedia and security: new challenges},
  pages={27--30},
  year={2001}
}

@article{song2024survey,
  title={A survey on Deep-Learning-based image steganography},
  author={Song, Bingbing and Wei, Ping and Wu, Sixing and Lin, Yu and Zhou, Wei},
  journal={Expert Systems with Applications},
  pages={124390},
  year={2024},
  publisher={Elsevier}
}

@article{zhang2020udh,
  title={Udh: Universal deep hiding for steganography, watermarking, and light field messaging},
  author={Zhang, Chaoning and Benz, Philipp and Karjauv, Adil and Sun, Geng and Kweon, In So},
  journal={Advances in Neural Information Processing Systems},
  volume={33},
  pages={10223--10234},
  year={2020}
}

@article{hayes2017generating,
  title={Generating steganographic images via adversarial training},
  author={Hayes, Jamie and Danezis, George},
  journal={Advances in neural information processing systems},
  volume={30},
  year={2017}
}

@article{baluja2017hiding,
  title={Hiding images in plain sight: Deep steganography},
  author={Baluja, Shumeet},
  journal={Advances in neural information processing systems},
  volume={30},
  year={2017}
}

@inproceedings{zhu2018hidden,
  title={Hidden: Hiding data with deep networks},
  author={Zhu, Jiren and Kaplan, Russell and Johnson, Justin and Fei-Fei, Li},
  booktitle={Proceedings of the European conference on computer vision (ECCV)},
  pages={657--672},
  year={2018}
}

@article{duan2019reversible,
  title={Reversible image steganography scheme based on a U-Net structure},
  author={Duan, Xintao and Jia, Kai and Li, Baoxia and Guo, Daidou and Zhang, En and Qin, Chuan},
  journal={Ieee Access},
  volume={7},
  pages={9314--9323},
  year={2019},
  publisher={IEEE}
}

@article{huang2022image,
  title={Image data hiding with multi-scale autoencoder network},
  author={Huang, Chen-Hsiu and Wu, Ja-Ling},
  journal={Electronic Imaging},
  volume={34},
  pages={1--6},
  year={2022},
  publisher={Society for Imaging Science and Technology}
}

@inproceedings{rahim2018end,
  title={End-to-end trained CNN encoder-decoder networks for image steganography},
  author={Rahim, Rafia and Nadeem, Shahroz and others},
  booktitle={Proceedings of the European conference on computer vision (ECCV) workshops},
  pages={0--0},
  year={2018}
}

@article{wu2018stegnet,
  title={Stegnet: Mega image steganography capacity with deep convolutional network},
  author={Wu, Pin and Yang, Yang and Li, Xiaoqiang},
  journal={Future Internet},
  volume={10},
  number={6},
  pages={54},
  year={2018},
  publisher={MDPI}
}

@inproceedings{lu2021large,
  title={Large-capacity image steganography based on invertible neural networks},
  author={Lu, Shao-Ping and Wang, Rong and Zhong, Tao and Rosin, Paul L},
  booktitle={Proceedings of the IEEE/CVF conference on computer vision and pattern recognition},
  pages={10816--10825},
  year={2021}
}

@inproceedings{he2015delving,
  title={Delving deep into rectifiers: Surpassing human-level performance on imagenet classification},
  author={He, Kaiming and Zhang, Xiangyu and Ren, Shaoqing and Sun, Jian},
  booktitle={Proceedings of the IEEE international conference on computer vision},
  pages={1026--1034},
  year={2015}
}

@inproceedings{xu2022robust,
  title={Robust invertible image steganography},
  author={Xu, Youmin and Mou, Chong and Hu, Yujie and Xie, Jingfen and Zhang, Jian},
  booktitle={Proceedings of the IEEE/CVF conference on computer vision and pattern recognition},
  pages={7875--7884},
  year={2022}
}

@article{guan2022deepmih,
  title={DeepMIH: Deep invertible network for multiple image hiding},
  author={Guan, Zhenyu and Jing, Junpeng and Deng, Xin and Xu, Mai and Jiang, Lai and Zhang, Zhou and Li, Yipeng},
  journal={IEEE Transactions on Pattern Analysis and Machine Intelligence},
  volume={45},
  number={1},
  pages={372--390},
  year={2022},
  publisher={IEEE}
}

@inproceedings{jing2021hinet,
  title={Hinet: Deep image hiding by invertible network},
  author={Jing, Junpeng and Deng, Xin and Xu, Mai and Wang, Jianyi and Guan, Zhenyu},
  booktitle={Proceedings of the IEEE/CVF international conference on computer vision},
  pages={4733--4742},
  year={2021}
}

@article{zhu2021destroying,
  title={Destroying robust steganography in online social networks},
  author={Zhu, Zhiying and Li, Sheng and Qian, Zhenxing and Zhang, Xinpeng},
  journal={Information Sciences},
  volume={581},
  pages={605--619},
  year={2021},
  publisher={Elsevier}
}

@article{zhu2022image,
  title={Image sanitization in online social networks: A general framework for breaking robust information hiding},
  author={Zhu, Zhiying and Wei, Ping and Qian, Zhenxing and Li, Sheng and Zhang, Xinpeng},
  journal={IEEE Transactions on Circuits and Systems for Video Technology},
  volume={33},
  number={6},
  pages={3017--3029},
  year={2022},
  publisher={IEEE}
}

@article{hu2024learning,
  title={Learning-based image steganography and watermarking: A survey},
  author={Hu, Kun and Wang, Mingpei and Ma, Xiaohui and Chen, Jia and Wang, Xiaochao and Wang, Xingjun},
  journal={Expert Systems with Applications},
  pages={123715},
  year={2024},
  publisher={Elsevier}
}

@article{liu2020coverless,
  title={Coverless image steganography based on DenseNet feature mapping},
  author={Liu, Qiang and Xiang, Xuyu and Qin, Jiaohua and Tan, Yun and Qiu, Yao},
  journal={EURASIP Journal on Image and Video Processing},
  volume={2020},
  pages={1--18},
  year={2020},
  publisher={Springer}
}

@article{yu2023cross,
  title={Cross: Diffusion model makes controllable, robust and secure image steganography},
  author={Yu, Jiwen and Zhang, Xuanyu and Xu, Youmin and Zhang, Jian},
  journal={Advances in Neural Information Processing Systems},
  volume={36},
  pages={80730--80743},
  year={2023}
}

@article{lecun1998gradient,
  title={Gradient-based learning applied to document recognition},
  author={LeCun, Yann and Bottou, L{\'e}on and Bengio, Yoshua and Haffner, Patrick},
  journal={Proceedings of the IEEE},
  volume={86},
  number={11},
  pages={2278--2324},
  year={1998},
  publisher={Ieee}
}

@inproceedings{kipf2017semi,
  title={Semi-Supervised Classification with Graph Convolutional Networks},
  author={Kipf, Thomas N and Welling, Max},
  booktitle={International Conference on Learning Representations},
  year={2017}
}

@article{hamilton2017inductive,
  title={Inductive representation learning on large graphs},
  author={Hamilton, Will and Ying, Zhitao and Leskovec, Jure},
  journal={Advances in neural information processing systems},
  volume={30},
  year={2017}
}

@article{han2022vision,
  title={Vision gnn: An image is worth graph of nodes},
  author={Han, Kai and Wang, Yunhe and Guo, Jianyuan and Tang, Yehui and Wu, Enhua},
  journal={Advances in neural information processing systems},
  volume={35},
  pages={8291--8303},
  year={2022}
}

@inproceedings{munir2023mobilevig,
  title={Mobilevig: Graph-based sparse attention for mobile vision applications},
  author={Munir, Mustafa and Avery, William and Marculescu, Radu},
  booktitle={Proceedings of the IEEE/CVF Conference on Computer Vision and Pattern Recognition},
  pages={2211--2219},
  year={2023}
}

@inproceedings{munir2024greedyvig,
  title={Greedyvig: Dynamic axial graph construction for efficient vision gnns},
  author={Munir, Mustafa and Avery, William and Rahman, Md Mostafijur and Marculescu, Radu},
  booktitle={Proceedings of the IEEE/CVF Conference on Computer Vision and Pattern Recognition},
  pages={6118--6127},
  year={2024}
}

@article{spadaro2024wignet,
  title={WiGNet: Windowed Vision Graph Neural Network},
  author={Spadaro, Gabriele and Grangetto, Marco and Fiandrotti, Attilio and Tartaglione, Enzo and Giraldo, Jhony H},
  journal={arXiv preprint arXiv:2410.00807},
  year={2024}
}

@inproceedings{dosovitskiy2020image,
  title={An Image is Worth 16x16 Words: Transformers for Image Recognition at Scale},
  author={Dosovitskiy, Alexey and Beyer, Lucas and Kolesnikov, Alexander and Weissenborn, Dirk and Zhai, Xiaohua and Unterthiner, Thomas and Dehghani, Mostafa and Minderer, Matthias and Heigold, G and Gelly, S and others},
  booktitle={International Conference on Learning Representations},
  year={2020}
}

@article{baluja2019hiding,
  title={Hiding images within images},
  author={Baluja, Shumeet},
  journal={IEEE transactions on pattern analysis and machine intelligence},
  volume={42},
  number={7},
  pages={1685--1697},
  year={2019},
  publisher={IEEE}
}

@inproceedings{imaizumi2014multibit,
  title={Multibit Embedding Algorithm for Steganography of Palette-Based Images},
  author={Imaizumi, Shoko and Ozawa, Kei},
  booktitle={Image and Video Technology: 6th Pacific-Rim Symposium, PSIVT 2013, Guanajuato, Mexico, October 28-November 1, 2013, Proceedings},
  volume={8333},
  pages={99},
  year={2014},
  organization={Springer}
}

@article{provos2003hide,
  title={Hide and seek: An introduction to steganography},
  author={Provos, Niels and Honeyman, Peter},
  journal={IEEE security \& privacy},
  volume={1},
  number={3},
  pages={32--44},
  year={2003},
  publisher={IEEE}
}

@inproceedings{ruanaidh1996phase,
  title={Phase watermarking of digital images},
  author={Ruanaidh, JJKO and Dowling, William J and Boland, Francis M},
  booktitle={Proceedings of 3rd IEEE International Conference on Image Processing},
  volume={3},
  pages={239--242},
  year={1996},
  organization={IEEE}
}

@article{hsu1999hidden,
  title={Hidden digital watermarks in images},
  author={Hsu, Chiou-Ting and Wu, Ja-Ling},
  journal={IEEE Transactions on image processing},
  volume={8},
  number={1},
  pages={58--68},
  year={1999},
  publisher={IEEE}
}

@article{barni2001improved,
  title={Improved wavelet-based watermarking through pixel-wise masking},
  author={Barni, Mauro and Bartolini, Franco and Piva, Alessandro},
  journal={IEEE transactions on image processing},
  volume={10},
  number={5},
  pages={783--791},
  year={2001},
  publisher={IEEE}
}

@inproceedings{tancik2020stegastamp,
  title={Stegastamp: Invisible hyperlinks in physical photographs},
  author={Tancik, Matthew and Mildenhall, Ben and Ng, Ren},
  booktitle={Proceedings of the IEEE/CVF conference on computer vision and pattern recognition},
  pages={2117--2126},
  year={2020}
}

@inproceedings{wengrowski2019light,
  title={Light field messaging with deep photographic steganography},
  author={Wengrowski, Eric and Dana, Kristin},
  booktitle={Proceedings of the IEEE/CVF conference on computer vision and pattern recognition},
  pages={1515--1524},
  year={2019}
}

@article{dinh2014nice,
  title={Nice: Non-linear independent components estimation},
  author={Dinh, Laurent and Krueger, David and Bengio, Yoshua},
  journal={arXiv preprint arXiv:1410.8516},
  year={2014}
}

@inproceedings{xiao2020invertible,
  title={Invertible image rescaling},
  author={Xiao, Mingqing and Zheng, Shuxin and Liu, Chang and Wang, Yaolong and He, Di and Ke, Guolin and Bian, Jiang and Lin, Zhouchen and Liu, Tie-Yan},
  booktitle={Computer Vision--ECCV 2020: 16th European Conference, Glasgow, UK, August 23--28, 2020, Proceedings, Part I 16},
  pages={126--144},
  year={2020},
  organization={Springer}
}

@article{ye2024pprsteg,
  title={PPRSteg: Printing and Photography Robust QR Code Steganography via Attention Flow-Based Model},
  author={Ye, Huayuan and Zhang, Shenzhuo and Jiang, Shiqi and Liao, Jing and Gu, Shuhang and Wang, Changbo and Li, Chenhui},
  journal={arXiv e-prints},
  pages={arXiv--2405},
  year={2024}
}

@article{zhang2019steganogan,
  title={SteganoGAN: High Capacity Image Steganography with GANs},
  author={Zhang, Kevin Alex and Cuesta-Infante, Alfredo and Xu, Lei and Veeramachaneni, Kalyan},
  journal={CoRR},
  year={2019}
}

@article{qin2019coverless,
  title={Coverless image steganography: a survey},
  author={Qin, Jiaohua and Luo, Yuanjing and Xiang, Xuyu and Tan, Yun and Huang, Huajun},
  journal={IEEE access},
  volume={7},
  pages={171372--171394},
  year={2019},
  publisher={IEEE}
}

@inproceedings{bruna2014spectral,
  title={Spectral networks and deep locally connected networks on graphs},
  author={Bruna, Joan and Zaremba, Wojciech and Szlam, Arthur and LeCun, Yann},
  booktitle={2nd International Conference on Learning Representations, ICLR 2014},
  year={2014}
}

@article{defferrard2016convolutional,
  title={Convolutional neural networks on graphs with fast localized spectral filtering},
  author={Defferrard, Micha{\"e}l and Bresson, Xavier and Vandergheynst, Pierre},
  journal={Advances in neural information processing systems},
  volume={29},
  year={2016}
}

@article{sen2008collective,
  title={Collective classification in network data},
  author={Sen, Prithviraj and Namata, Galileo and Bilgic, Mustafa and Getoor, Lise and Galligher, Brian and Eliassi-Rad, Tina},
  journal={AI magazine},
  volume={29},
  number={3},
  pages={93--93},
  year={2008}
}

@article{wale2008comparison,
  title={Comparison of descriptor spaces for chemical compound retrieval and classification},
  author={Wale, Nikil and Watson, Ian A and Karypis, George},
  journal={Knowledge and Information Systems},
  volume={14},
  pages={347--375},
  year={2008},
  publisher={Springer}
}

@inproceedings{xu2017scene,
  title={Scene graph generation by iterative message passing},
  author={Xu, Danfei and Zhu, Yuke and Choy, Christopher B and Fei-Fei, Li},
  booktitle={Proceedings of the IEEE conference on computer vision and pattern recognition},
  pages={5410--5419},
  year={2017}
}

@inproceedings{giraldo2022hypergraph,
  title={Hypergraph convolutional networks for weakly-supervised semantic segmentation},
  author={Giraldo, Jhony H and Scarrica, Vincenzo and Staiano, Antonino and Camastra, Francesco and Bouwmans, Thierry},
  booktitle={2022 IEEE international conference on image processing (ICIP)},
  pages={16--20},
  year={2022},
  organization={IEEE}
}

@article{russakovsky2015imagenet,
  title={Imagenet large scale visual recognition challenge},
  author={Russakovsky, Olga and Deng, Jia and Su, Hao and Krause, Jonathan and Satheesh, Sanjeev and Ma, Sean and Huang, Zhiheng and Karpathy, Andrej and Khosla, Aditya and Bernstein, Michael and others},
  journal={International journal of computer vision},
  volume={115},
  pages={211--252},
  year={2015},
  publisher={Springer}
}

@inproceedings{lin2014microsoft,
  title={Microsoft coco: Common objects in context},
  author={Lin, Tsung-Yi and Maire, Michael and Belongie, Serge and Hays, James and Perona, Pietro and Ramanan, Deva and Doll{\'a}r, Piotr and Zitnick, C Lawrence},
  booktitle={Computer vision--ECCV 2014: 13th European conference, zurich, Switzerland, September 6-12, 2014, proceedings, part v 13},
  pages={740--755},
  year={2014},
  organization={Springer}
}

@inproceedings{zhang2018unreasonable,
  title={The unreasonable effectiveness of deep features as a perceptual metric},
  author={Zhang, Richard and Isola, Phillip and Efros, Alexei A and Shechtman, Eli and Wang, Oliver},
  booktitle={Proceedings of the IEEE conference on computer vision and pattern recognition},
  pages={586--595},
  year={2018}
}

@article{wang2004image,
  title={Image quality assessment: from error visibility to structural similarity},
  author={Wang, Zhou and Bovik, Alan C and Sheikh, Hamid R and Simoncelli, Eero P},
  journal={IEEE transactions on image processing},
  volume={13},
  number={4},
  pages={600--612},
  year={2004},
  publisher={IEEE}
}

@article{boehm2014stegexpose,
  title={StegExpose-A Tool for Detecting LSB Steganography},
  author={Boehm, Benedikt},
  journal={arXiv preprint arXiv:1410.6656},
  year={2014}
}

@article{boroumand2018deep,
  title={Deep residual network for steganalysis of digital images},
  author={Boroumand, Mehdi and Chen, Mo and Fridrich, Jessica},
  journal={IEEE Transactions on Information Forensics and Security},
  volume={14},
  number={5},
  pages={1181--1193},
  year={2018},
  publisher={IEEE}
}

@article{wang2022pvt,
  title={Pvt v2: Improved baselines with pyramid vision transformer},
  author={Wang, Wenhai and Xie, Enze and Li, Xiang and Fan, Deng-Ping and Song, Kaitao and Liang, Ding and Lu, Tong and Luo, Ping and Shao, Ling},
  journal={Computational visual media},
  volume={8},
  number={3},
  pages={415--424},
  year={2022},
  publisher={Springer}
}

@inproceedings{islammuch,
  title={How much Position Information Do Convolutional Neural Networks Encode?},
  author={Islam, Md Amirul and Jia, Sen and Bruce, Neil DB},
  booktitle={International Conference on Learning Representations}
}

@inproceedings{chuconditional,
  title={Conditional Positional Encodings for Vision Transformers},
  author={Chu, Xiangxiang and Tian, Zhi and Zhang, Bo and Wang, Xinlong and Shen, Chunhua},
  booktitle={The Eleventh International Conference on Learning Representations}
}

@inproceedings{liu2021swin,
  title={Swin transformer: Hierarchical vision transformer using shifted windows},
  author={Liu, Ze and Lin, Yutong and Cao, Yue and Hu, Han and Wei, Yixuan and Zhang, Zheng and Lin, Stephen and Guo, Baining},
  booktitle={Proceedings of the IEEE/CVF international conference on computer vision},
  pages={10012--10022},
  year={2021}
}

@inproceedings{li2024purified,
  title={Purified and unified steganographic network},
  author={Li, Guobiao and Li, Sheng and Luo, Zicong and Qian, Zhenxing and Zhang, Xinpeng},
  booktitle={Proceedings of the IEEE/CVF conference on computer vision and pattern recognition},
  pages={27569--27578},
  year={2024}
}

@inproceedings{giraldo2023trade,
  title={On the trade-off between over-smoothing and over-squashing in deep graph neural networks},
  author={Giraldo, Jhony H and Skianis, Konstantinos and Bouwmans, Thierry and Malliaros, Fragkiskos D},
  booktitle={Proceedings of the 32nd ACM international conference on information and knowledge management},
  pages={566--576},
  year={2023}
}

@inproceedings{zhang2018residual,
  title={Residual dense network for image super-resolution},
  author={Zhang, Yulun and Tian, Yapeng and Kong, Yu and Zhong, Bineng and Fu, Yun},
  booktitle={Proceedings of the IEEE conference on computer vision and pattern recognition},
  pages={2472--2481},
  year={2018}
}

@inproceedings{zeiler2010deconvolutional,
  title={Deconvolutional networks},
  author={Zeiler, Matthew D and Krishnan, Dilip and Taylor, Graham W and Fergus, Rob},
  booktitle={2010 IEEE Computer Society Conference on computer vision and pattern recognition},
  pages={2528--2535},
  year={2010},
  organization={IEEE}
}

@inproceedings{agustsson2017ntire,
  title={Ntire 2017 challenge on single image super-resolution: Dataset and study},
  author={Agustsson, Eirikur and Timofte, Radu},
  booktitle={Proceedings of the IEEE conference on computer vision and pattern recognition workshops},
  pages={126--135},
  year={2017}
}

@article{kingma2014adam,
  title={Adam: A method for stochastic optimization},
  author={Kingma, Diederik P and Ba, Jimmy},
  journal={arXiv preprint arXiv:1412.6980},
  year={2014}
}

@article{wang2019dynamic,
  title={Dynamic graph cnn for learning on point clouds},
  author={Wang, Yue and Sun, Yongbin and Liu, Ziwei and Sarma, Sanjay E and Bronstein, Michael M and Solomon, Justin M},
  journal={ACM Transactions on Graphics (tog)},
  volume={38},
  number={5},
  pages={1--12},
  year={2019},
  publisher={Acm New York, NY, USA}
}

@inproceedings{velivckovic2018graph,
  title={Graph Attention Networks},
  author={Veli{\v{c}}kovi{\'c}, Petar and Cucurull, Guillem and Casanova, Arantxa and Romero, Adriana and Li{\`o}, Pietro and Bengio, Yoshua},
  booktitle={International Conference on Learning Representations},
  year={2018}
}

@article{bianchi2023expressive,
  title={The expressive power of pooling in graph neural networks},
  author={Bianchi, Filippo Maria and Lachi, Veronica},
  journal={Advances in neural information processing systems},
  volume={36},
  pages={71603--71618},
  year={2023}
}

@article{yang2021graphformers,
  title={Graphformers: Gnn-nested transformers for representation learning on textual graph},
  author={Yang, Junhan and Liu, Zheng and Xiao, Shitao and Li, Chaozhuo and Lian, Defu and Agrawal, Sanjay and Singh, Amit and Sun, Guangzhong and Xie, Xing},
  journal={Advances in Neural Information Processing Systems},
  volume={34},
  pages={28798--28810},
  year={2021}
}
}

\end{document}